\documentclass[journal]{IEEEtran}

\usepackage[utf8]{inputenc}
\usepackage{blindtext}
\usepackage{graphicx}
\usepackage{booktabs}
\usepackage{amssymb}
\usepackage{mathtools}
\usepackage{tabularx}
\usepackage{accents}
\usepackage{todonotes}
\usepackage{url}
\usepackage{hyperref}
\usepackage[square,sort,comma,numbers]{natbib}
\usepackage{multirow}
\usepackage{booktabs}
\usepackage{enumitem}
\usepackage{todonotes}
\usepackage{subcaption}
\usepackage{bbm}

\begin{document}

\title{Disentangle, align and fuse for multimodal and semi-supervised image segmentation}

\author{Agisilaos~Chartsias,
        Giorgos~Papanastasiou,
        Chengjia~Wang, 
        Scott~Semple, 
        David~E.~Newby, 
        Rohan~Dharmakumar,
        Sotirios~A.~Tsaftaris % ~\IEEEmembership{Senior Member,~IEEE}

\thanks{This work was supported by US National Institutes of Health (1R01HL136578-01) and used resources from the Edinburgh Compute and Data Facility. S.A. Tsaftaris is also supported by a Canon Medical / Royal Academy of Engineering Research Chair (RCSRF1819\textbackslash8\textbackslash25). (Corresponding author: S.A. Tsaftaris)}
\thanks{A.~Chartsias and S.A.~Tsaftaris are with the School of Engineering, The University of Edinburgh, Edinburgh, EH9~3JL, UK. 
S.A.~Tsaftaris is also with The Alan Turing Institute, London, NW1~2DB, UK  (email: Agis.Chartsias@ed.ac.uk, S.Tsaftaris@ed.ac.uk).
G.~Papanastasiou, C.~Wang, S.~Semple, and D.E.~Newby are with the Edinburgh Imaging Facility QMRI and with the Centre for Cardiovascular Science, Edinburgh, EH16~4TJ, UK (email: G.Papanas@ed.ac.uk, chengjia.wang@ed.ac.uk, scott.semple@ed.ac.uk, D.E.Newby@ed.ac.uk). R.~Dharmakumar is with the Cedars Sinai Medical Center Los Angeles CA 90048, USA. (email: Rohan.Dharmakumar@cshs.org)}
}

% \markboth{IEEE TRANSACTIONS ON MEDICAL IMAGING,~Vol.?, No.?, November~2019}%
% {Shell \MakeLowercase{\textit{et al.}}: Bare Demo of IEEEtran.cls for Journals}

\maketitle

\begin{abstract}

Magnetic resonance (MR) protocols rely on several sequences to assess pathology and organ status properly. Despite advances in image analysis, we tend to treat each sequence, here termed modality, in isolation. Taking advantage of the common information shared between modalities (an organ's anatomy) is beneficial for multi-modality processing and learning. However, we must overcome inherent anatomical misregistrations and disparities in signal intensity across the modalities to obtain this benefit. 
We present a method that offers improved segmentation accuracy of the modality of interest (over a single input model), by learning to leverage information present in other modalities, even if few (semi-supervised) or no (unsupervised) annotations are available for this specific modality.
Core to our method is learning a disentangled decomposition into anatomical and imaging factors. Shared anatomical factors from the different inputs are jointly processed and fused to extract more accurate segmentation masks. Image misregistrations are corrected with a Spatial Transformer Network, which non-linearly aligns the anatomical factors. 
The imaging factor captures signal intensity characteristics across different modality data and is used for image reconstruction, enabling semi-supervised learning. Temporal and slice pairing between inputs are learned dynamically. 
We demonstrate applications in Late Gadolinium Enhanced (LGE) and Blood Oxygenation Level Dependent (BOLD) cardiac segmentation, as well as in T2 abdominal segmentation. 
Code is available at \url{https://github.com/vios-s/multimodal_segmentation}.
\end{abstract}


% \IEEEpeerreviewmaketitle

\section{Introduction}

\IEEEPARstart{I}n medical imaging multiple acquisitions of the same subject capture complementary information. Within Magnetic Resonance (MR), different pulse sequences (modalities) attenuate different tissue characteristics to identify anatomical and functional information and produce images of different contrasts.
Automatic segmentation of such data remains important and relies on accurate annotations, which however can be expensive to acquire, thus resulting in scarce or imperfectly annotated datasets. To overcome this, many solutions have been proposed for example with the use of unlabelled data~\cite{tajbakhsh2020embracing}.

We approach this problem with a method that leverages data from a secondary modality and is based on disentangled representations. Our method is designed to address challenges of multimodal data that include differences in signal intensities, lack of annotated data, and anatomical and temporal misalignments due to varying spatial resolutions or due to moving organs, as in the case of dynamic heart imaging.

\begin{figure}[t!]
\centering
\includegraphics[width=\linewidth]{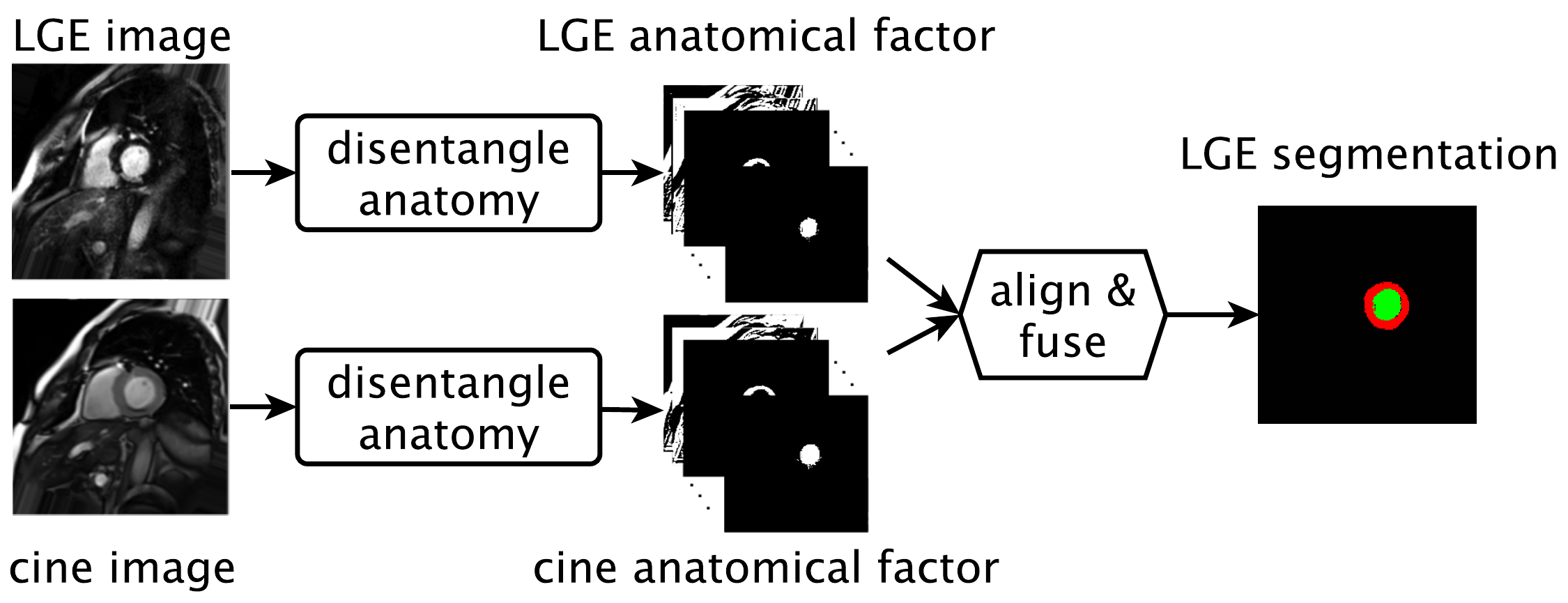}
\caption{DAFNet schematic in an LGE segmentation exemplar task using LGE and cine-MR inputs. Firstly, \textit{disentangled} anatomical factors are extracted from the inputs. Then, they are \textit{aligned} (with a Spatial Transformer~\cite{jaderberg2015spatial}) and combined to a \textit{fused} anatomy that infers the final LGE segmentation. Our approach can use multi-input (multimodal) data at training and inference. The latter is extremely useful when training with \textit{zero annotations} for an input and also in removing outliers.}
\label{fig:figure1}
\end{figure}

Multimodal learning allows capturing of information present in one modality (e.g. the anatomy) for use in another modality that has greater contrast.
As a motivating example, myocardial segmentation in LGE is challenging, since LGE nulls myocardial signal to accentuate signal originating from myocardial infarction. In fact, in clinical practice, analysis of LGE is typically combined with cine-MR~\cite{kim2009cardiovascular}.

A naive way to propagate knowledge between modalities would be co-registration. This has been successful in the brain (see Section~\ref{sec:related_multimodal}) but precise multimodal registration needs modality independent metrics~\cite{sotiras2013} and the brain remains static within an imaging session, whereas the heart is constantly moving. 
Furthermore, multimodal cardiac data are often inconsistent both in the number of images (different slices, cardiac phases, and perhaps more penalising 
resolution differences, e.g. slice thickness) and in the number of annotations. 
In addition, some sequences are static (LGE) and others 
dynamic~(cine-MR). These challenges necessitate solutions that alleviate misregistrations and can pair input images. 

\subsection{Overview of the proposed approach}

We propose a mechanism that can propagate knowledge for segmentation and can learn with and without annotations using a reconstruction objective as self-supervision. Moreover, our approach co-registers data within an anatomical representation space, becoming thus robust to variations in imaging contrast. 
Our 2D approach, Disentangle Align and Fuse Network (DAFNet), achieves the above by mapping multimodal images of the same subject into \textit{disentangled} anatomy and modality factors.\footnote{In computer vision these are typically referred to as content and style factors, respectively~\cite{huang2018multimodal}. However, as we detail in Section~\ref{sec:related_disentangled} in medical applications such type of disentanglement has more stringent requirements.} 
At inference time, as Figure~\ref{fig:figure1} shows, DAFNet fuses the disentangled anatomy factors to combine multimodal information and extract segmentation masks.

Anatomy factors are represented as categorical feature maps with each category corresponding to input pixels that are ideally spatially similar and hence belong to the same anatomical part. This promotes semantic consistency and helps learn spatial correspondences between anatomical parts from different modalities.
Modality factors encode pixel intensities in a smooth multivariate Gaussian manifold as per the Variational Autoencoder (VAE)~\cite{kingma2013auto}. 
Anatomy factors are used by a segmentation network to obtain masks, whereas a decoder re-entangles both factors to achieve image reconstruction. 

Learning a disentangled representation is encouraged by minimising the information capacity of each factor respectively: thresholding the anatomy factors, which results in binarised discrete feature channels, prevents the storage of image intensity information as continuous values, which impairs disentanglement~\cite{chartsias2019disentangled}, whereas the variational objective minimises information in the modality factors~\cite{alemi2016deep}. Disentanglement is also influenced by the decoder design, either through inductive biases (see discussion in Section~\ref{sec:decoding} and evaluation in Section~\ref{sec:exp_disentanglement}), or through learning constraints (see cross-modal decoding of one anatomy in the modality of another in Section~\ref{sec:decoding}, similar to~\cite{huang2018multimodal}).

However, a disentangled representation with modality invariant anatomy factors is not enough for multimodal learning. A multimodal representation also requires similar anatomy factors across modalities, such that each feature channel corresponds to the same anatomical region for any modality. This is achieved by weight sharing in the anatomy encoders, as well as by shared segmentation and decoder networks. 
These constraints implicitly create common anatomy semantics, which are essential when no labels, but only images, are available for one of the modalities.
In this case we project all images to the common anatomical space, where a single segmentation network is trained with supervision only on the annotated modalities. 
Finally, when input data are \textit{not} paired a new loss term in the cost function selects the most ``informative'' multimodal pairs by comparing anatomy factors.\footnote{We improve our preliminary work \cite{chartsias2019multimodal} as follows: (1) we reduce model parameters and encourage common anatomy semantics by employing weight sharing in the anatomy encoders; (2) we introduce a cost that reduces the need for expert pairing of multimodal inputs; (3) we design and evaluate another decoder; (4) we propose the use of distance correlation to assess disentanglement; (5) we generalise to non-cardiac datasets.}

Our contributions are summarised as follows:
\begin{enumerate}
    \item We propose a 2D method for learning disentangled representations of anatomy and modality factors in multimodal medical images for  segmentation.
    \item We demonstrate the importance of semantic anatomy factors, achieved through the model design, since they allow learning registration and fusion operators.
    \item We propose a loss term in the cost function that enables the selection of the most ``informative'' multimodal pairs.
    \item We demonstrate our method's robustness over other approaches with extensive experiments  on several datasets, in cardiac MRI and abdominal segmentation.
    \item We show that our model works both on single-modal and multimodal inference, and that it outperforms other variants when trained with different amount (semi-supervised) or zero annotations for one of the modalities.
    \item We discuss decoder designs using FiLM~\cite{perez2017film} and SPADE~\cite{park2019semantic} respectively and evaluate disentanglement by estimating the dependence between anatomy and modality factors with distance correlation~\cite{szekely2007measuring}.
\end{enumerate}

\section{Related work} \label{sec:related_work}
Multimodal machine learning is a research area that involves diverse modalities.
While in computer vision modalities might refer to heterogeneous sources of information, such as text and images, here, as common in the medical domain, we restrict to different image acquisitions.
We consider multimodal learning as combining information from different images, present at training or more critically at inference time. Using image translation or domain adaptation as augmentation strategy to reconcile lack of annotated data also enables training, although not inference, with multimodal images (we term this multimodal inference).\footnote{For completeness we mention few recent methods. Image translation is proposed with cycle consistency~\cite{campello2019combining, huo2018synseg, zhang2018translating, tao2019segmentation} or disentanglement~\cite{chen2019unsupervised} losses. 
In domain adaptation, multimodal images are related with different augmentations~\cite{ly2019style}, histogram 
matching~\cite{liu2019automatic} or adversarial losses~\cite{Chen2019SynergisticIA}.}

We review work on disentangled representations, the main focus of our method, and also on multimodal medical imaging, split in spatially registered or unregistered inputs. We further discuss differences between other methods and ours regarding the representation semantics and the approach to fusion, in particular when having misaligned inputs and limited annotations.
We highlight that currently no work exists that is able to simultaneously achieve multimodal fusion from unregistered data for segmentation, be robust to the number of training annotations, and be applied to single or multimodal inference. 

\subsection{Disentangled representation learning} \label{sec:related_disentangled}
Disentangling content from style for style transfer is gaining popularity in computer vision with many examples, such as~\cite{huang2018multimodal}. In medical imaging, disentangled representations have been used for semi-supervised cardiac segmentation~\cite{chartsias2018factorised, chartsias2019disentangled}, multi-task learning~\cite{chartsias2019disentangled, meng2019representation}, lung nodule synthesis~\cite{liu2018decompose}, and multimodal registration~\cite{qin2019unsupervised}. Disentangling multimodal images has also been used for liver segmentation with domain adaptation~\cite{yang2019unsupervised} albeit without information fusion, and for brain tumour segmentation~\cite{chen2019robust} although using registered images.

For anatomical features to be useful in clinical tasks, they need to be semantic and quantifiable. This is not guaranteed in disentanglement techniques used in style transfer~\cite{huang2018multimodal} and in medical segmentation~\cite{yang2019unsupervised, chen2019robust} that do not impose restrictions on the content features. Semantic representations have recently been pursued in computer vision in the form of feature masks~\cite{ma2018exemplar} or by learning geometry with landmarks~\cite{wu2019transgaga}. 
Instead, we disentangle quantifiable anatomical features so that they are useful in segmentation, whereas interpretability is promoted with explicit design constraints (Section~\ref{sec:encoding}).

\subsection{Multimodal learning with registered images} \label{sec:related_multimodal}

Early work on multimodal deep learning concatenated co-registered images in different input channels to improve MR brain segmentation~\cite{havaei2017brain}. Robustness to missing modalities~\cite{joyce2017robust} was achieved with encoders per modality that mapped images to modality invariant features. Common features with multiple encoders was also proposed for cross-modal classification~\cite{van2019learning}.  

Another aspect of multimodal learning is information fusion. Most commonly, fusion is performed on latent features~\cite{havaei2017brain, joyce2017robust}, although fusion at multiple levels can be achieved with densely connected layers~\cite{dolz2018hyperdense} to exploit multi-scale correlations. 
Furthermore, cross-modal convolutions are used to weigh each modality's contribution~\cite{tseng2017joint}. Finally, attention modules and residual learning focus on specific regions for brain MRI segmentation \cite{chen2018mmfnet}.
In contrast to the above, we take advantage of the strictly defined anatomy factors and use a $max$ fusion operator to combine all distinct features.

\subsection{Multimodal learning with unregistered images} \label{sec:related_multimodal_unregistered}

Unfortunately, misalignment between multimodal images remains common. Correcting small misalignments is possible with a Spatial Transformer Network (STN)~\cite{jaderberg2015spatial} applied on features~\cite{joyce2017robust}. Alternatively, encoder-decoder setups can be trained with multimodal data to learn common features. An exploration of different setups~\cite{valindria2018multi} showed that separate encoders and decoders sharing their last and first layer respectively, achieve the highest performance. 

In cardiac image analysis approaches are limited. Multiple inputs can be combined by adapting segmentation masks with contour models~\cite{liu2019multi, wei2011myocardial}. Alternatively, reducing the field of view (to patch level) and ensembling (fusing results from several atlases) can alleviate the effect of errors~\cite{zhuang2016multi}. Recently, simultaneous segmentation and registration of multimodal cardiac MR, models the joint distribution as Multivariate Mixtures~\cite{zhuang2018multivariate}.
Multimodal images can also be used as samples of the same data distribution to form an expanded dataset~\cite{wang2019skunet}. 
Finally, multimodal registration, although susceptible to errors, can create ``noisy'' labels~\cite{roth2019cardiac}.
Our method is the first to learn 
a suitable representation for segmentation, co-registration, 
and information fusion, \textit{but} in a semi-supervised setting. 

\begin{figure*}[t!]
\centering
\includegraphics[width=\linewidth]{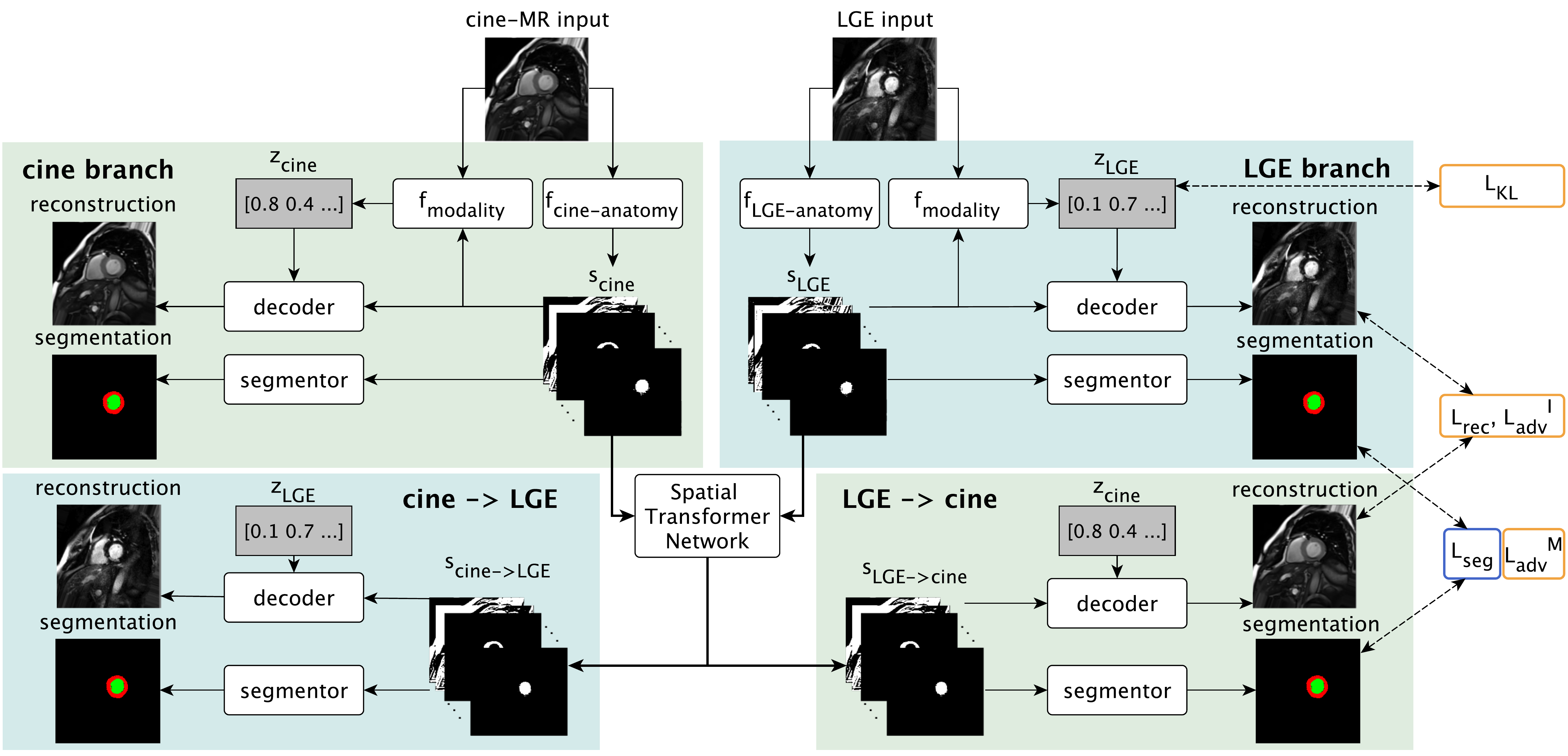}
\caption{Training schematic with cine-MR and LGE input images. Each input image is disentangled into anatomical and modality factors. With a Spatial Transformer Network the deformation branches (lower parts) enable cross-modal synthesis and segmentation by deforming the anatomy factors $s_{cine}$ and $s_{LGE}$.
Losses are indicated on the right and are also symmetrically applied to the cine-MR branch outputs. Yellow or blue outlines indicate if a loss is used when training with zero or full supervision, respectively. $L_{rec}^z$ is not shown. See text for definitions. }
\label{fig:mmsdnet_train}
\end{figure*}

\section{Methodology} \label{sec:methodology}
Here we describe DAFNet,
a multi-component 2D model for multimodal and semi-supervised segmentation that is illustrated (at training time) in Figure~\ref{fig:mmsdnet_train}. Initially, input images are encoded into anatomy and modality factors (Section~\ref{sec:encoding}). Anatomy factors are aligned (with a STN) and fused (Section~\ref{sec:alignment_fusion}), and participate in segmentation losses (Section~\ref{sec:segmentation}). Training also employs image reconstruction (Section~\ref{sec:decoding}) and modality reconstruction losses (Section~\ref{sec:modality_reconstruction}). Finally, a multimodal pairing loss allows to dynamically learn how to pair image sources (Section~\ref{sec:pair_cost}). Below we detail the individual components, as well as the 
employed cost functions. 

\subsection{Encoding into anatomy and modality factors} \label{sec:encoding}

\begin{figure}
\centering
\includegraphics[width=\linewidth]{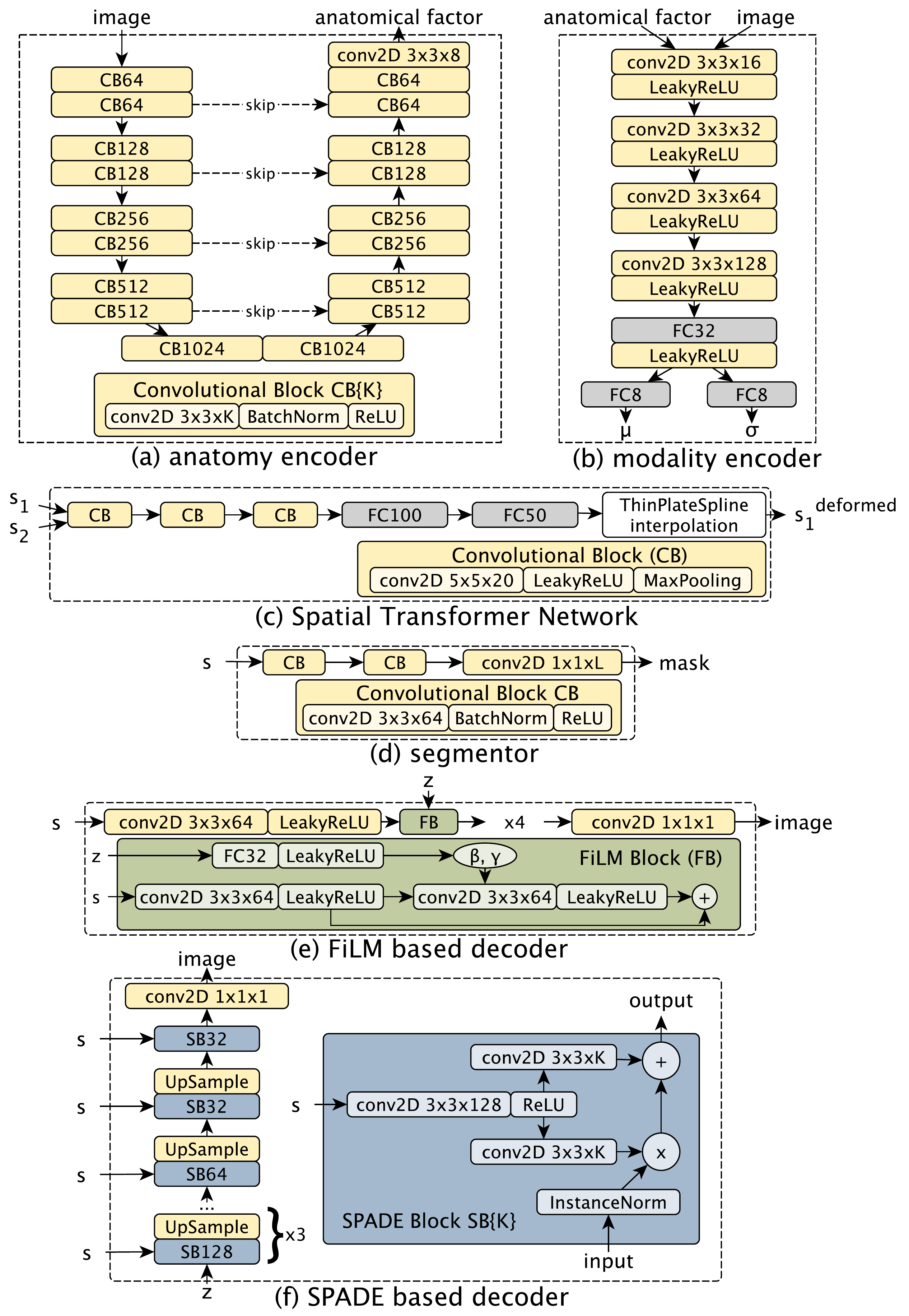}
\caption{Architecture diagrams of the individual components. (a) Encoder extracts anatomy factors; (b) Modality encoder extracts parameters $\mu$, $\sigma$ of a Gaussian distribution. The modality factor is a sample from this distribution; (c) The segmentation network produces a mask given an anatomy factor; (d) The Spatial Transformer Network receives two anatomy factors and produces the 2D co-ordinates of 25 control points used for interpolation; (e) Decoder architecture based on FiLM~\cite{perez2017film}; (f) Decoder architecture based on SPADE~\cite{park2019semantic}.}
\label{fig:architectures}
\end{figure}

Given modality $i \in \left\lbrace M_1, \ldots, M_n \right\rbrace$ with samples $x_i \in X_i$, where $X_i \subset {\rm I\!R}^{H \times W}$ is the set of images, and $H$ and $W$ are the height and width respectively, the encoding process achieves a disentanglement of anatomy and modality factors. 
Anatomy factors $s_i$ are tensors produced by encoders dedicated to each modality $i$: $s_i=f_{anatomy}(x_i|\theta_i)$, where $\theta_i$ are the encoder parameters. The encoder architecture is modelled after the U-Net~\cite{ronneberger2015u} and is shown in Figure~\ref{fig:architectures}a. To reduce model parameters and encourage a common anatomical representation among the multimodal data, we employ weight sharing in the decoder of each U-Net. Thus, parameters $\theta_i$ are split into individual parameters $\phi_i$ of the encoding path, and shared parameters $\rho$ of the decoding path: $s_i=f_{anatomy}(x_i|\phi_i, \rho)$.

\begin{figure}[t]
\centering
\includegraphics[width=\linewidth]{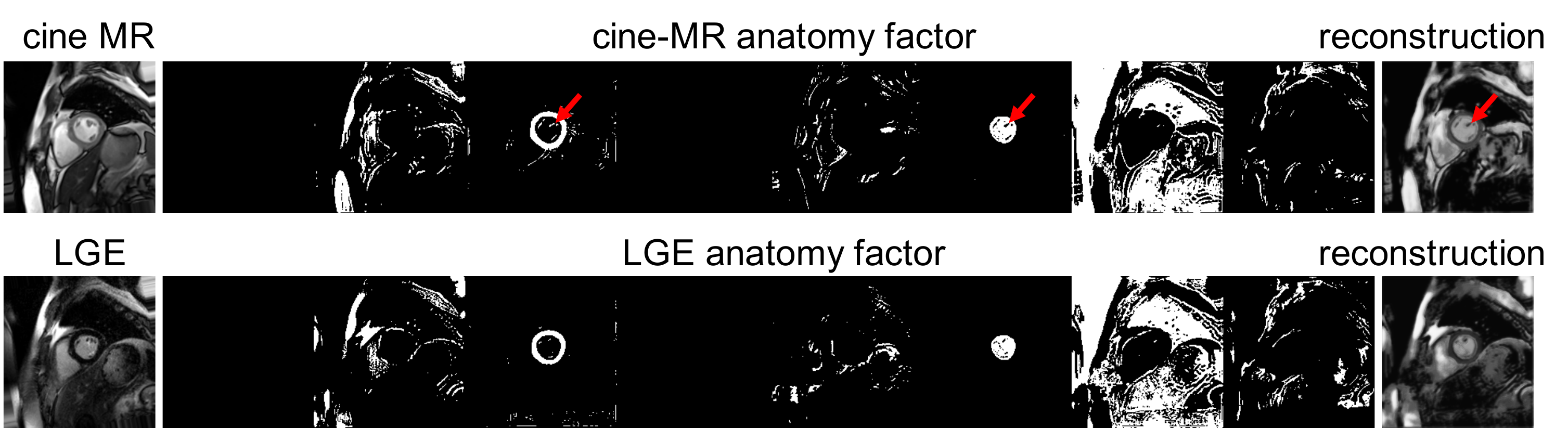}
\caption{Cine and LGE anatomy factors. Observe that the same anatomical regions appear in the same channels. Also, encoding papillary muscles (marked in red arrows) in myocardium and left ventricle channels allows their reconstruction (which would be impossible if these channels were segmentations).}
\label{fig:anatomical_factors}
\end{figure}

An anatomy factor is represented as a binary tensor that is also a one-hot encoding in the channel dimension. A binary anatomy factor cannot store different pixel intensities of an image as continuous values and this promotes the factorisation process~\cite{chartsias2019disentangled}. Furthermore, this one-hot encoding encourages a particular image region to appear in a single channel. More formally, $s_i \in \{0,1\}^{H \times W \times C}$, s.t. $\sum_{c=1}^{C} s_i^{h,w,c} = 1$ $\forall h \in \{1, \ldots, H\}, w \in \{1, \ldots, W\}$. Two example anatomy factors produced by a cine and a LGE image can be seen in Figure~\ref{fig:anatomical_factors}. 

Although binary, the anatomy factors are not segmentations. Instead they must be useful in segmentation and image decoding tasks described in the following sections.\footnote{In fact, segmentations cannot be directly used in the decoding task when these contain multiple anatomical structures, as this causes tension between the segmentation and decoding tasks. As an example, consider papillary muscles, which here are part of left ventricle segmentations. Synthetic images cannot contain papillary muscles unless they are separated in the anatomy encoding (Figure~\ref{fig:anatomical_factors}). We break this tension with a segmentation network, which combines different channels of an anatomy factor to produce a segmentation.}

\noindent \textbf{Divergence loss $L_{KL}$:} Our model also decomposes images into a modality factor $z_i$, which aims to capture intensity and texture characteristics of the input data. The modality factors $z_i \in Z \coloneqq {\rm I\!R}^{n_z}$ are vectors produced by a single stochastic encoder, which, given an image sample $x_i$ and its anatomy factor $s_i$, learns a probability distribution $q(z_i|x_i,s_i)$. In order to encourage a smooth space and also minimise the encoded information~\cite{alemi2016deep}, the posterior distribution is optimised to follow a multivariate Gaussian prior, $p(z_i)=\mathcal{N}(\mathbf{0},\mathbf{I})$, using the $KL-$divergence and the re-parameterisation trick~\cite{kingma2013auto}:
\begin{equation*}
    L_{KL} = \int p(z_i)\log \frac{p(z_i)}{q(z_i|x_i,s_i)}dx_{i}ds_i.
\end{equation*}
The modality encoder is shown in Figure~\ref{fig:architectures}b and predicts the mean and standard deviation of a Gaussian, which are used to draw a random sample vector $z_i=f_{modality}(x_i, s_i)$.

\subsection{Alignment and fusion of the anatomy factors} \label{sec:alignment_fusion}

Following factor encoding, two anatomy factors $s_i$ and $s_j$ of modalities $i$ and $j$ respectively, are aligned using non-linear registration. Given a grid of $5\times 5$ control points, a STN (architecture in Figure~\ref{fig:architectures}c) first predicts the grid's offsets. Then, thin plate spline interpolates the surface passing through the control points to register $s_j$ with $s_i$. The result of alignment, $s_i^{deformed}$, is a deformed anatomy factor corresponding to $s_j$ and vice versa ($s_j^{deformed}$ corresponds to $s_i$). We optimise STN with gradients in image space (self-supervised cost of Section~\ref{sec:decoding}) and with the segmentation cost of Section~\ref{sec:segmentation}, since we aim to align segmentation masks.\footnote{We avoid direct comparison of $s_i$ and $s_j^{deformed}$, since they are binary and thus different small deformations might generate the same error.}

During inference the deformed anatomies are combined to produce a fused representation, containing all unique and shared features that are present in the constituent anatomy factors. Since they are spatially aligned, a pixel wise operation such as the pixel-wise max is able to preserve all encoded features. More formally, $s_i^{fused}=\max(s_i, s_j^{deformed})$ and $s_j^{fused}=\max(s_j, s_i^{deformed})$. One benefit of max-fusion is that it is invariant to the number of inputs and is therefore directly applicable in cases with more than two modalities.

\subsection{Segmentation} \label{sec:segmentation}

Given an anatomy factor $s_i$, a simple neural network (architecture in Figure~\ref{fig:architectures}d) infers a corresponding segmentation mask $m_i=h(s_i)$, s.t. $m_i \in M_i \coloneqq \{0, 1\}^{H \times W \times L}$, where $M_i$ is the set of masks of modality $i$ and $L$ is the number of segmentation classes. The segmentation network is common for all modalities, is applied to the deformed and fused anatomies of Section~\ref{sec:alignment_fusion} and is optimised as follows.

\noindent \textbf{Supervised loss $L_{sup}$:} Given a set of images paired with masks $(x_i, m_i)$, a supervised cost is defined as a weighted sum of the differentiable Dice loss and Cross Entropy~(CE):
\begin{equation*}
L_{sup}=\alpha (1-Dice(h(s_i), m_i)) + \beta CE(h(s_i), m_i),
\end{equation*}
where $\alpha$ and $\beta$ control the contribution of each loss. 
The cross entropy and differentiable Dice are respectively defined as:
\begin{equation*}
    CE(h(s_i), m_i)= - \sum_l(m_{h,w,l}\log(p_{h,w,c})),
\end{equation*} 
\begin{equation*}
    Dice(h(s_i), m_i)=2 \times \left[ \frac{\sum_{h,w,c} (h(s_{i_{h,w,c}}) \times m_{i_{h,w,c}})}{\sum_{h,w,c} (h(s_{i_{h,w,c}}) + m_{i_{h,w,c}})} \right],
\end{equation*}
where $h$, $w$, and $c$ refer to the height, width and channel, and $p_{h,w,l}$ is the probability for a pixel belonging to class $l$.

\noindent \textbf{Adversarial loss $L^M_{adv}$:} An unsupervised segmentation cost is also defined with a mask discriminator $D_M$, modelled after LS-GAN~\cite{mao2017effectiveness}. The adversarial objective given real masks sampled from all modalities $m \sim M_i, i \in \{1,2,\ldots n\}$ is:
\begin{equation*}
    L^M_{adv} = D_M(h(s_i))^2 + (D_M(m)-1)^2,
\end{equation*}
where the discriminator is adversarially trained against the segmentation network. The discriminator's architecture consists of 4 convolutional layers followed by LeakyReLU and a final single neuron layer
and uses Spectral Normalisation~\cite{miyato2018spectral} to stabilise training.
In both segmentation costs, the anatomy factors $s_i$ come either from the input images directly or are the result of the alignment step of a secondary $j$ modality: $s_i \in \{f_{anatomy}(x_i|\theta_i), s_j^{deformed}\}$. In the latter case, the gradients produced by the segmentation cost are back-propagated to the STN module to learn its parameters.\footnote{We omit the use of $s_j^{fused}$ as input to the segmentation network to avoid backpropagating gradients both to the STN and the $j^{th}$ anatomy encoder, which might result in the STN not achieving a good convergence.}

Segmentation losses help learn better anatomy factors that separate the anatomies of interest in respective channels (see myocardium and left ventricle in Figure~\ref{fig:anatomical_factors}). If supervision is not available for modality $i$, training is performed with the adversarial loss $L_{adv}^M$. In addition, the deformed anatomy factor $s_i^{deformed}$ participates in the supervised segmentation of another modality $j$. The latter encourages anatomy factors to be similarly represented across modalities and thus enables unsupervised segmentation of modality $i$.

\subsection{Decoding} \label{sec:decoding}

The anatomy factors are further decoded into an output image of a style dictated by a modality factor $z_i$: $y_i=g(s_i, z_i)$. This factor entanglement can be performed with different decoders, which also indirectly influence the type of disentanglement, or in other words the type of information captured by the anatomy and modality factors. We investigate two decoder architectures based on FiLM~\cite{perez2017film} and SPADE~\cite{park2019semantic}. 

The input of the \textbf{FiLM}-based decoder (Figure~\ref{fig:architectures}e) is the anatomy factors, which, after a series of convolutions, are conditioned by $z$ samples. These are used to predict a scale and an offset parameter $\gamma \in {\rm I\!R}^{C}$ and $\beta \in {\rm I\!R}^{C}$, which modulate each intermediate feature map $F \in {\rm I\!R}^{H \times W \times C}$, where $H$, $W$ and $C$ are the height, width and number of channels respectively: $FiLM(F|\gamma, \beta)=F \odot \gamma + \beta$. 

We also consider a \textbf{SPADE}-based decoder (Figure~\ref{fig:architectures}f), which has been demonstrated to generate texture details on synthetic images given segmentation masks. The input to this decoder is a $z$ sample, which is processed by a series of convolutional layers and conditioned by the anatomy factor defining the output ``shape''. An Instance Normalisation layer with parameters $\mu$ and $\sigma$ is firstly applied to a feature map $F \in {\rm I\!R}^{H \times W \times C}$, which is then modulated by tensors $\mathbf{\Gamma}$ and $\mathbf{B}$ (same size as $F$)  $SPADE(F|\mathbf{\Gamma}, \mathbf{B})=\mathbf{\Gamma} \odot \frac{F-\mu}{\sigma}+\mathbf{B}$.

\noindent \textbf{Self-supervised costs $L_{rec}$ and $L^{I,i}_{adv}$:} The decoders are trained to reconstruct the input with the $\ell_1$ loss:
\begin{equation*}
    L_{rec}=\|x_i-g(s_i, z_i)\|_1.
\end{equation*}
In addition, synthesis of realistic images is encouraged with an image discriminator $D_{I,i}$ for a modality $i$, modelled after LS-GAN. This defines an adversarial loss:
\begin{equation*}
    L^{I}_{adv} = D_{I,i}(g(s_i, z_i))^2 + (D_{I,i}(x_i)-1)^2.
\end{equation*}
As in the segmentation case, $s_i$ is an encoding of image $x_i$ or a deformed encoding of another image $x_j$: $s_i \in \{f_{anatomy}(x_i|\theta_i), s_j^{deformed}\}$. When $s_i=f_{anatomy}(x_i|\theta_i)$, the model acts as an auto-encoder. This is critical to allow the use of non-annotated images and thus enable semi-supervised learning. In the case where $s_i=s_j^{deformed}$, the backpropagated gradients train the STN module and also aid the factorisation process (since the ``style'' of the output image $g(s_j^{deformed}, z_i)$, as specified by $z_i$, corresponds to modality $i$ and not $j$).

The decoder also participates in the following loss that promotes disentanglement.

\subsection{Reconstruction of the modality factor loss $L^z_{rec}$} \label{sec:modality_reconstruction}

In order to encourage disentanglement, and also avoid posterior collapse of the modality factor, we reconstruct the modality factor of a synthetic image. This prevents the decoder from ignoring the $z$-factors and only use the anatomy factors for image synthesis. We minimise the following loss:
\begin{equation*}
    L^z_{rec}=\|z-f_{modality}(y, f_{anatomy}(y))\|_1,
\end{equation*}
where $z$ is a sample from a unit Gaussian and $y$ is the synthetic image produced by this $z$ sample. Encouraging the use of modality factors by the decoder is further achieved by cross-reconstructing a deformed anatomy in a modality dictated by the corresponding $z$-factor, as described in Section~\ref{sec:decoding}.

\subsection{Non-expert pairing} \label{sec:pair_cost}
Better multimodal fusion and STN registration will be achieved by multimodal image pairs $\{x_i, x_j\}$ that are more similar in terms of their spatial and temporal positions.  In cases where the multimodal images are not expertly paired, we can automatically measure anatomical similarities with an optional cost, which directly compares the anatomy factors, and ``selects'' only the most informative image pairs. 

During training, and given an image $x_i$ and a set of $k$ candidate images from modality $j$: $\{x_j^1, x_j^2, \ldots, x_j^k\}$, the multimodal segmentation and reconstruction losses for a sample $x_i$ are weighted accordingly by $k$ weights, s.t. $\sum_{l=1}^k w_l=1$: 
\begin{equation*}
    L_{sup}=\sum^k_{l=1} w_l L_{sup}(m_i, m_l), 
    L_{rec}=\sum^k_{l=1} w_l L_{rec}(x_i, y_l),
\end{equation*}
where $m_l=h(f_{anatomy}(x^l_j))$, and $y_l=g(s_j^{deformed, l}, z_i)$. 
By weighting the loss functions, the STN module does not need to learn deformations for all pairs, as well as prevents it from trying to match slices with different anatomical content.

Due to the semantics of the anatomy factors, and the fact that they are categorical, we can directly evaluate their overlap in terms of the Dice score. The Dice for each pair, becomes the input to a small neural network of two fully connected layers that outputs the weights, and is similar to the temperature scaling technique proposed for calibrating classification outputs~\cite{guo2017calibration}. At inference time, the most accurate $m_i$ segmentation is produced from the weighted sum of the fusion with different slices $s_j^{fused,1}, s_j^{fused,2}, \ldots s_j^{fused,k}$:
\begin{equation*}
    m_i=w_1 h(s_j^{fused,1}) + w_2 h(s_j^{fused,2}) + \ldots + w_k h(s_j^{fused,k}).
\end{equation*}
This optional weighting of the cost function is only used in unpaired data, and as shown in experiment~\ref{sec:exp_pairs} converges to the same result as manual pairing.

\section{Experimental setup} \label{sec:experimental_setup}

\subsection{Training details} 

The model is trained with a multi-component loss function, 
$L=  0.1\cdot L_{KL}+ 10\cdot L_{sup} + L_{adv}^M + L_{rec} + L_{adv}^I + L^z_{rec}.$
A higher $L_{sup}$ weight encourages separation of  segmentation classes. A reduced $L_{KL}$ weight prevents posterior collapse, in which the decoder ignores the $z$ factor; however, an even lower $L_{KL}$ would not well approximate the Gaussian prior leading to a non-smooth intensity manifold.
Number of $s$ channels and $z$ dimensions are set to $C$=8 and $n_z$=8 respectively, as in~\cite{chartsias2019disentangled}.

Code is written in Keras (\url{https://keras.io}) and is available at \url{https://github.com/vios-s/multimodal_segmentation} along with pseudocode of the training process. We train with Adam (learning rate $10^{-4}$) and use Stochastic Weight Averaging \cite{izmailov2018averaging} to reliably compare methods. Training takes approximately 12 hours on a Titan-X GPU and inference of one image takes 50 ms.
Quantitative evaluation is performed on 3-fold cross-validation, where the training, validation and test sets correspond to 70\%, 15\% and 15\% of the data volumes.

\subsection{Data} \label{sec:data}
Experiments use three multimodal datasets of a source and a target modality that have been rescaled to $[-1, 1]$.

\begin{enumerate}
    \item For LGE segmentation, we use cine-MR and LGE data of 28 patients~\cite{stirrat2017ferumoxytol}, acquired at Edinburgh Royal Infirmary (\textbf{ERI}), with spatial resolution $1.562$mm$^2$/pixel, and slice thickness $9mm$. End diastolic myocardial contours are provided. The image size is $192\times192$ pixels. The number of segmented images is 358 (for each of cine-MR, LGE).
    \item To evaluate robustness on different medical data, we use abdominal T1-dual inphase and T2-SPIR images from \textbf{CHAOS}~\cite{kavur2020chaos} dataset, for T2 segmentation. Images of 20 subjects with liver, kidneys and spleen segmentations are acquired by a 1.5T Philips MRI scanner, which produces 12-bit DICOM images of $256\times256$ resolution. We resample to an x-y spacing of 1.89mm, and crop to $192\times192$ pixels. In total, there are 1594 images.
    \item Finally, we evaluate BOLD segmentation with a dataset (shorthand \textbf{BOLD}) of cine-MR and CP-BOLD images~\cite{tsaftaris2013detecting} of 10 (mechanically ventilated) subjects, with an in-plane resolution of $1.25mm\times1.25mm$, acquired at baseline and severe ischemia (inflicted as controllable stenosis of the left-anterior descending coronary artery (LAD) on a 1.5T Espree (Siemens Healthineers). The image acquisition is at short axis view, covering the mid-ventricle, and is performed using cine-MR and a flow and motion compensated CP-BOLD sequence, where each sequence is applied one after the other in the protocol in separate breath-holds. The pixel resolution is $192 \times 114$. In total there are 129 cine-MR and 264 CP-BOLD images with expert segmentations from all cardiac phases.
\end{enumerate}

\subsection{Baseline and benchmark methods} \label{sec:benchmarks}

We consider the following baselines, which assume source masks being available at inference time. If predicted masks were used e.g. the result of a  U-Net, an additional confounder would be introduced. Thus, we report numbers with ground truth masks for a bias-free estimate, which albeit is elevated.
\begin{enumerate}
    \item A lower bound computes the Dice score between real masks of two modalities, which is also a measure of misalignment of the multimodal data. This is referred to as ``\textbf{copy}'' and can be used for segmenting a target modality without annotations from the target modality. 
    \item This lower bound can be improved after registering the multimodal images and applying the registration field to the source masks. The deformation field is calculated by affine registration using mutual information, followed by symmetric diffeomorphic using cross-correlation \cite{Tustison2015}.
    This is referred to as ``\textbf{register}'' baseline and can also be used without annotations of the target modality. ``Copy'' and ``register'' are common in clinical evaluation.
    \item Finally, we implement a version of a non-coupled \textbf{active contour} model akin to the one in~\cite{wei2011myocardial}. We initialised the contour using the ``copy'' above. Then via grid search, we found optimal contour length, smoothness, and stepping hyperparameters as: for ERI [0.5, 0.15, 0.7], BOLD [0.01, 0.15, 0.7] and CHAOS [0.5, 0.15, 0.7], respectively.
    \end{enumerate}

We also consider the following deep learning benchmarks.
\begin{enumerate}
    \item As a supervised benchmark we train a UNet on annotated data of the target modality and refer to it as \textbf{UNet-single}. 
    We further re-train a UNet on mixed training data of all modalities to evaluate its capability of concurrently handling multimodal data and refer to it as \textbf{UNet-multi}.
    \item We train SDNet~\cite{chartsias2019disentangled} with full or semi supervision on data of the target modality and refer to it as \textbf{SDNet-single}. We also train SDNet by mixing multimodal data, as demonstrated in~\cite{chartsias2019disentangled}, and refer to it as \textbf{SDNet-multi}.
    
    \item We get two final benchmarks by training MUNIT~\cite{huang2018multimodal} for image translation. The first uses MUNIT to translate images from source to target modality~\cite{chen2019unsupervised}, and the second translates multimodal images to a domain invariant space~\cite{yang2019unsupervised}. In both cases, segmentation is performed \textit{post-hoc} with a UNet on the combined data. We refer to these approaches as \textbf{Translation} and \textbf{DADR} respectively.
    \item Finally, we implement \textbf{DualStream} \cite{valindria2018multi}, the most recent Deep Learning based method for handling multimodal data which does not require registered data.
\end{enumerate}

For all benchmarks we use the architectures of the original papers, resulting in a comparable number of parameters. For UNet, SDNet, MUNIT, DualStream and DAFNet these are approximately in millions 35, 35, 26, 56 and 52, respectively.

\section{Experiments and discussion} \label{sec:experiments}

Sections \ref{sec:multi_single_segmentation} and \ref{sec:semisupervised} present segmentation results, assuming a source modality that always contains annotations during training. The \textit{source} modality is cine-MR for ERI and BOLD datasets, and T1 for CHAOS. The \textit{target} modality is LGE, BOLD and T2 for ERI, BOLD and CHAOS, respectively. Unless explicitly specified, DAFNet uses a FiLM-based decoder, and we report test Dice of the fused anatomies.
We evaluate the effects of \textit{input pairing} (Section~\ref{sec:exp_pairs}), \textit{registration} (Section~\ref{sec:exp_registration}), and a \textit{SPADE-based} decoder  (Section~\ref{sec:exp_decoder}). Then, Section \ref{sec:exp_disentanglement} evaluates \textit{disentanglement} of each decoder design. Where appropriate, bold font denotes the best (on average) method and an asterisk (*) denotes statistical significance of paired t-tests ($p<0.05$ assessed via permutations) comparing with the second best (to avoid multiple comparisons).  

\begin{table}[t!]
\begin{center}
\caption{Segmentation results on three datasets when full (100\%) or zero (0\%) target modality annotations are available. For each dataset we show results on the target modality assuming the other is the source (and vice versa). Single input-output models cannot be trained with no annotations and are marked with $n/a$. We omit results marked with $-$, since training of these methods did not converge.}
\label{table:mult_inputs}
\begin{tabularx}{\linewidth}{p{.11\linewidth}p{.045\linewidth}p{.045\linewidth}p{.077\linewidth}|p{.03\linewidth}p{.045\linewidth}p{.045\linewidth}p{.045\linewidth}p{.03\linewidth}p{.025\linewidth}}
\toprule
\multirow{3}{*}{Methods} & \multirow{3}{*}{Train} & \multirow{3}{*}{Test} & & \multicolumn{6}{c}{\textit{100\% target annotations}} \\
& & & Masks & \multicolumn{2}{c}{ERI} & \multicolumn{2}{c}{BOLD} & \multicolumn{2}{c}{CHAOS} \\
& & & in test & LGE & cine & BOLD & cine & T2 & T1 \\
\midrule
copy       & --     & multi   & Yes & $67_{06}$ & $67_{06}$ & $80_{01}$ & $80_{01}$ & $71_{10}$ & $71_{10}$ \\
register   & --     & multi   & Yes & $68_{07}$ & $67_{05}$ & $81_{04}$ & $84_{05}$ & $70_{07}$ & $73_{05}$ \\
AC         & --     & multi   & Yes & $66_{15}$ & $66_{13}$ & $68_{02}$ & $72_{05}$ & $65_{22}$ & $65_{22}$ \\
UNet       & single & single  & No  & $78_{04}$ & $85_{08}$ & $\textbf{91}_{01}$ & $89_{01}$ & $\textbf{85}_{17}$ & $86_{05}$ \\
SDNet      & single & single  & No  & $80_{03}$ & $84_{09}$ & $89_{03}$ & $88_{04}$ & $83_{16}$ & $85_{08}$ \\
UNet       & multi  & single  & No  & $81_{03}$ & $83_{08}$ & $89_{03}$ & $88_{02}$ & $\textbf{85}_{15}$ & $\textbf{88}_{03}$ \\
SDNet      & multi  & single  & No  & $80_{05}$ & $\textbf{86}_{05}$ & $89_{02}$ & $87_{03}$ & $\textbf{85}_{11}$ & $\textbf{88}_{01}$ \\
DualStream & multi  & single  & No  & $80_{06}$ & $\textbf{86}_{09}$ & $89_{09}$ & $88_{02}$ & $\textbf{85}_{16}$ & $85_{09}$ \\
Translation & multi & single  & No  & $79_{06}$ & $84_{05}$ & $83_{06}$ & $88_{02}$ & $83_{09}$ & $87_{06}$ \\
DADR       & multi  & single  & No  & $79_{05}$ & $83_{06}$ & $88_{04}$ & $86_{02}$ & $84_{16}$ & $72_{22}$ \\
DAFNet    & multi  & single  & No  & $\textbf{82}_{03}$ & $\textbf{86}_{02}$ & $88_{01}$ & $\textbf{91}_{02}$ & $83_{17}$ & $\textbf{88}_{01}$ \\
DAFNet    & multi  & multi   & No  & $\textbf{82}_{03}$ & $84_{02}$ & $\textbf{91}_{01}$ & $\textbf{91}_{01}$ & $\textbf{85}^*_{05}$ & $87_{01}$ \\
\midrule
\midrule
\multirow{3}{*}{Methods} & \multirow{3}{*}{Train} & \multirow{3}{*}{Test} & & \multicolumn{6}{c}{\textit{0\% target annotations}} \\
& & & Masks & \multicolumn{2}{c}{ERI} & \multicolumn{2}{c}{BOLD} & \multicolumn{2}{c}{CHAOS} \\
& & & in test & LGE & cine & BOLD & cine & T2 & T1 \\
\midrule
copy       & --     & multi   & Yes & $67_{06}$ & $67_{06}$ & $80_{01}$ & $80_{01}$ & $71_{10}$ & $71_{10}$ \\
register   & --     & multi   & Yes & $68_{07}$ & $67_{05}$ & $81_{04}$ & $84_{05}$ & $70_{07}$ & $73_{05}$ \\
AC         & n/a     & multi   & Yes & $66_{15}$ & $66_{13}$ & $68_{02}$ & $72_{05}$ & $65_{22}$ & $65_{22}$ \\
UNet       & single & single  & No  & n/a        & n/a        & n/a        & n/a        & n/a        & n/a \\
SDNet      & single & single  & No & n/a         & n/a        & n/a        & n/a        & n/a        & n/a \\
UNet       & multi  & single  & No & $38_{23}$  & $68_{12}$ & $68_{23}$ & $85_{05}$ & -- & -- \\
SDNet      & multi  & single  & No & $61_{18}$  & $73_{07}$ & $80_{03}$ & $85_{03}$ & $51_{09}$ & $63_{13}$ \\
DualStream & multi  & single  & No & $38_{23}$  & $68_{12}$ & $68_{23}$ & $85_{05}$ & -- & -- \\
Translation & multi & single  & No & $37_{23}$  & $61_{13}$ & $61_{10}$ & $74_{07}$ & -- & $45_{11}$ \\
DADR       & multi  & single  & No & $46_{19}$  & $63_{13}$ & $68_{11}$ & $85_{01}$ & -- & $49_{17}$ \\
DAFNet    & multi  & single  & No & $72_{06}$  & $\textbf{78}_{05}$ & $78_{02}$ & $82_{03}$ & $72_{12}$ & $\textbf{74}_{06}$ \\
DAFNet    & multi  & multi   & No & $\textbf{74}^*_{04}$ & $76_{04}$ & $\textbf{85}^*_{03}$ & $\textbf{86}_{02}$ & $\textbf{74}^*_{03}$ & $71_{06}$ \\
\bottomrule
\end{tabularx}
\end{center}
\end{table}

\subsection{Multimodal segmentation: full and zero supervision setting} \label{sec:multi_single_segmentation}

The prime contribution of our work is the ability to learn and infer in a multimodal setting.
Thus, we first demonstrate that multiple inputs at training and inference time benefit segmentation. Table~\ref{table:mult_inputs} presents test Dice scores on three datasets for DAFNet and the benchmarks of Section~\ref{sec:benchmarks}. Two setups are evaluated, assuming either that annotations are available for the target modality or not.

In the 100\% case, training with multiple inputs improves accuracy, even when multimodal data simply constitute an augmented dataset. When segmenting the target modality, the usage of multiple inputs at inference time by DAFNet, obtains similar Dice as other benchmarks, but considerably reduces the standard deviation, such as in CHAOS from 11\% to 5\%.

In the 0\% case the (learned) benchmark methods do not produce accurate target segmentations for all datasets. As expected, models trained only on the source modality learn modality-specific features, and as such cannot generalise to the unseen target modality. DAFNet on the other hand, consistently maintains a better average and smaller variance by leveraging information from the source modality. This is due to the aligning (in an embedding sense) of the multimodal representations in the anatomy space, which allows the shared segmentor trained with supervision on the source, to also segment the target modality with ``zero'' supervised examples.

We then exchange source and target and report the cine-MR and T1 Dice by training new models where appropriate. 
The CP-BOLD sequence is very similar to cine, showing anatomy, but has elevated T2 contrast (the BOLD effect) \cite{tsaftaris2013detecting}. In addition, these data are acquired in controlled experiments, with the cine-MR and BOLD images acquired one after the other in the protocol. Thus, for the BOLD dataset all methods perform well in the 0\% case of cine-MR segmentation, with multiple inputs further improving DAFNet performance.
On the contrary, segmenting LGE (and T2) is more difficult. The Dice is overall lower than the cine-MR in the single-output DAFNet with the difference being bigger in the 0\% case. As a result, this hurts the multi-output cine-MR Dice.
This is expected since the benefit of multimodal segmentation comes when one modality is easier to segment.\footnote{Indeed, cine-MR is designed to show anatomical information, whereas LGE to highlight infarcted myocardium.} Therefore LGE benefits when considering cine-MR images, but the contrary would only be beneficial in cine-MR reconstruction problems, e.g. in the presence of motion artefacts~\cite{oksuz2019detection}.

\begin{table}[t!]
\centering
\caption{Segmentation results of LGE, BOLD and T2, when training with a varying amount of annotations for ERI, BOLD, and CHAOS datasets respectively.}
\label{table:semisupervised}
\begin{tabularx}{\linewidth}{l|ccc}
\toprule
\multicolumn{4}{c}{\textit{ERI: Target LGE Dice (Source cine)}} \\ 
\midrule
Method & ~~~~~~50\%~~~~~~ & ~~~~~~25\%~~~~~~ & ~~~~~~12.5\%~~~~~~ \\
\midrule
copy        & $67_{06}$ & $67_{06}$ & $67_{06}$ \\
register    & $68_{07}$ & $68_{07}$ & $68_{07}$ \\
AC & $66_{15}$ & $66_{15}$ & $66_{15}$ \\
UNet-single & $76_{12}$ & $66_{14}$ & $51_{21}$ \\
SDNet-single & $76_{04}$ & $69_{09}$ & $54_{18}$ \\
UNet-multi   & $76_{08}$ & $67_{11}$ & $50_{19}$ \\
SDNet-multi & $76_{04}$ & $73_{07}$ & $64_{19}$ \\
DualStream  & $76_{03}$ & $61_{13}$ & $44_{23}$ \\
Translation & $75_{07}$ & $67_{14}$ & $62_{14}$ \\
DADR & $77_{05}$ & $66_{11}$ & $57_{19}$ \\
DAFNet     & $\textbf{78}_{04}^*$ & $\textbf{76}^*_{05}$ & $\textbf{74}^*_{05}$ \\
\midrule
\midrule
\multicolumn{4}{c}{\textit{BOLD: Target BOLD Dice (Source cine)}} \\ 
\midrule
Method & 50\% & 25\% & 12.5\% \\
\midrule
copy        & $80_{01}$ & $80_{01}$ & $80_{01}$ \\
register    & $81_{04}$ & $81_{04}$ & $81_{04}$ \\
AC          & $68_{02}$ & $68_{02}$ & $68_{02}$ \\
UNet-single & $79_{17}$ & $59_{27}$ & $49_{29}$ \\
SDNet-single & $84_{03}$ & $68_{17}$ & $64_{14}$ \\
UNet-multi   & $\textbf{87}_{03}$ & $75_{17}$ & $72_{13}$ \\
SDNet-multi & $86_{07}$ & $85_{03}$ & $80_{03}$ \\
DualStream  & $86_{01}$ & $58_{26}$ & $49_{28}$ \\
Translation & $84_{02}$ & $79_{06}$ & $47_{26}$ \\
DADR        & $\textbf{87}_{02}$ & $79_{01}$ & $71_{15}$ \\
DAFNet     & $\textbf{87}_{01}$ & $\textbf{86}_{03}$ & $\textbf{85}^*_{03}$ \\
\midrule
\midrule
\multicolumn{4}{c}{\textit{CHAOS: Target T2 Dice (Source T1)}} \\ 
\midrule
Method & 50\% & 25\% & 12.5\% \\
\midrule
copy        & $71_{10}$ & $71_{10}$ & $71_{10}$ \\
register    & $70_{07}$ & $70_{07}$ & $70_{07}$ \\
AC          & $65_{22}$ & $65_{22}$ & $65_{22}$ \\
UNet-single & $80_{17}$ & $76_{15}$ & $72_{17}$ \\
SDNet-single & $82_{14}$ & $77_{16}$ & $75_{14}$ \\
UNet-multi   & $\textbf{84}_{15}$ & $79_{16}$ & $75_{16}$ \\
SDNet-multi & $\textbf{84}_{11}$ & $80_{13}$ & $78_{09}$ \\
DualStream  & $81_{19}$ & $78_{16}$ & $75_{16}$ \\
Translation & $81_{07}$ & $75_{11}$ & $70_{10}$ \\
DADR & $\textbf{84}_{11}$ & $77_{14}$ & $74_{11}$ \\
DAFNet     & $\textbf{84}_{05}$ & $\textbf{82}_{03}$ & $\textbf{79}_{05}^*$ \\
\bottomrule
\end{tabularx}
\end{table}

\subsection{Semi-supervised segmentation} \label{sec:semisupervised}

\begin{figure*}[t!]
\centering
\includegraphics[width=\linewidth]{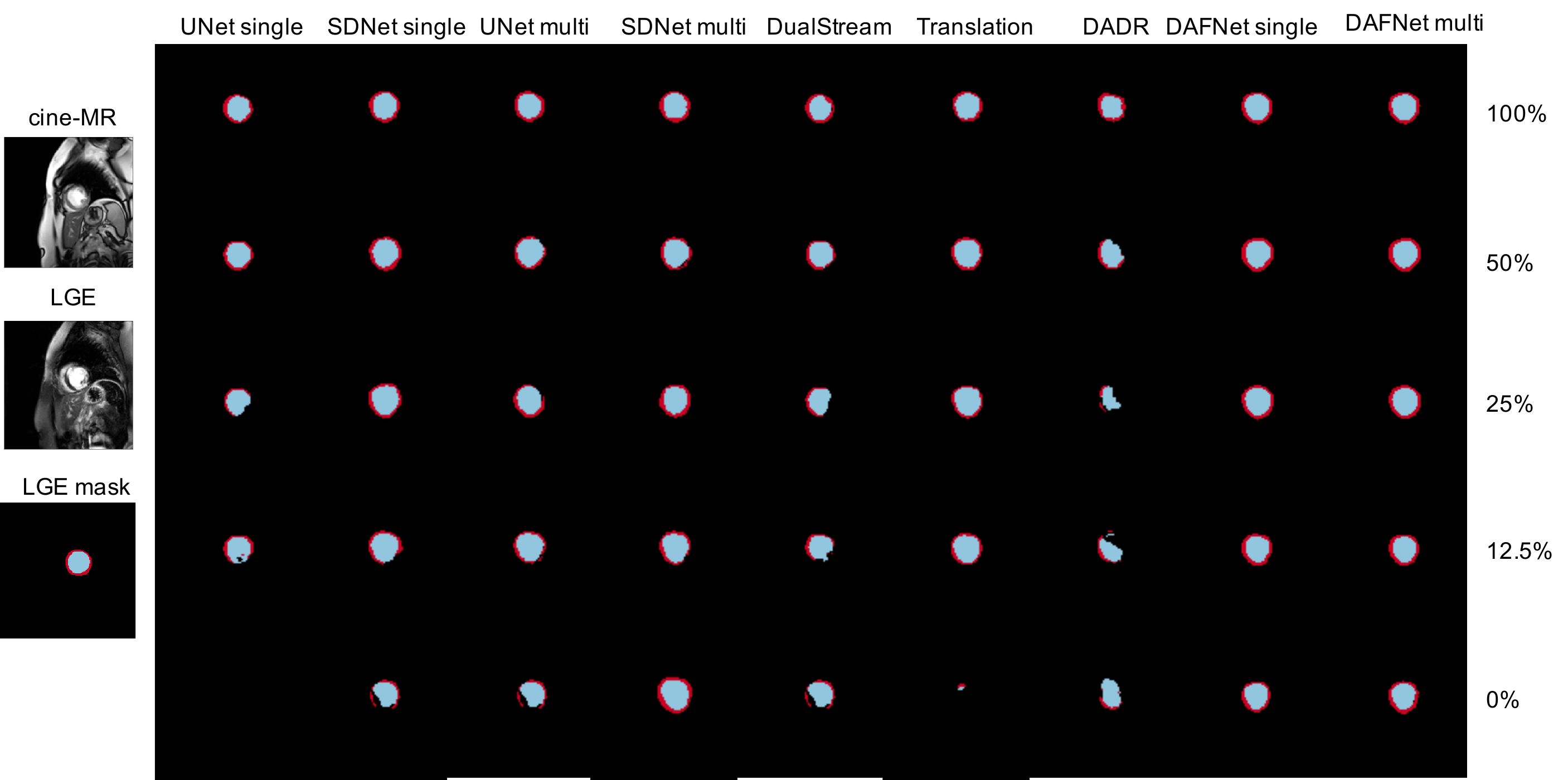}
\caption{Panel of LGE segmentation examples from ERI dataset, obtained with different amount of LGE annotations.}
\label{fig:semisupervised_segmentations}
\end{figure*}

Here we evaluate the sensitivity of all methods on different amounts of ground truth annotations available during training. 
Table~\ref{table:semisupervised} presents the average (across all labels) cross-validation test set Dice score. Exemplar test results are shown in Figure~\ref{fig:semisupervised_segmentations}. The number of images for both source and target modalities are fixed, but the amount of target annotations varies. Sampling the amount of annotations is performed on a subject-level, to avoid having a mixture of annotated and non-annotated images of the same subject in the training set. The DAFNet results correspond to using multiple inputs at inference time.

Average Dice for all methods is comparable when the number of annotations is high, although DAFNet achieves the lowest variance. With a reducing number of annotations, the performance of the competing methods is also reduced with a simultaneous increase in the variance. DAFNet maintains good results and robustness to edge cases, as evidenced by the small variance achieved throughout all setups.

\subsection{Effect of pair matching} \label{sec:exp_pairs}
The results of Sections \ref{sec:multi_single_segmentation} and \ref{sec:semisupervised} correspond to expertly paired multimodal inputs. Here, we evaluate the sensitivity of DAFNet on unpaired multimodal images, as well as the effect of the automated pairing cost proposed in Section~\ref{sec:pair_cost}.

We randomly shuffle the multimodal pairs by two positions, with the shuffled pairs differing up to two spatial slices within a 3D volume.\footnote{Similar results can be obtained by shuffling the different cardiac phases in the cine-MR temporal stack.}
We measure the LGE segmentation Dice score on ERI data when using 100\% and 0\% LGE annotations.
We thus compare our automated method with expert pairing (upper bound) and a random shuffle (lower bound). Table~\ref{table:mult_inputs_nonexpert} presents the results of copy method, as well as of DAFNet evaluated with both cine-MR and LGE inputs.

\begin{figure}[t!]
\centering
\includegraphics[width=\linewidth]{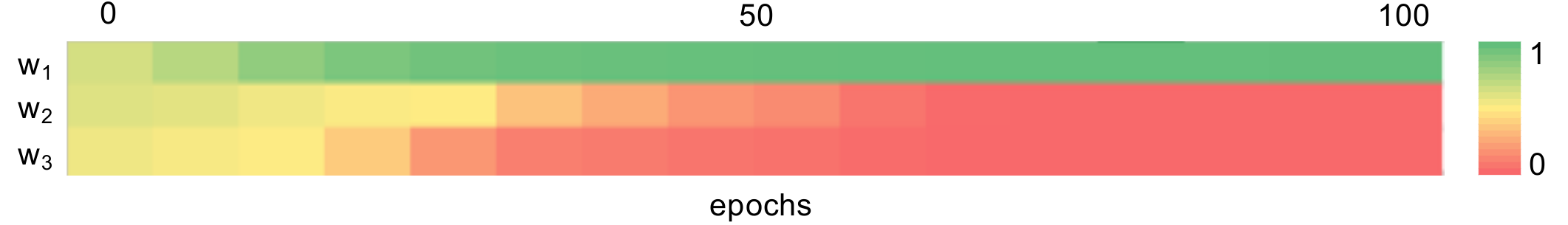}
\caption{Evolution of weights $w_j$ across epochs. Weights are used as a measure of similarity between each candidate multimodal pair. For more details see text.}
\label{fig:pair_weights}
\end{figure}

Shuffling the multimodal pairs decreases the copy performance considerably.  In both cases, automated matching of candidate pairs based on the semantics of the anatomy factors proves effective in ignoring distant slices (in the volume), with results very closely approaching the ones achieved by expert pairing.  To show how our model learns appropriate weights, the evolution of weights across training epochs is shown in Figure~\ref{fig:pair_weights}, in which $w_1$ corresponds to the closest pair converged to a probability of one early on in training. 

During inference, a ``soft'' segmentation mask is produced as a weighted sum between each weight with its corresponding mask. However, this converges to using the prediction of the ``closest'' pair, as evidenced by Figure~\ref{fig:pair_weights}.

\begin{table}[t!]
\begin{center}
\caption{LGE segmentation results when multimodal images are not expertly paired.}
\label{table:mult_inputs_nonexpert}
\begin{tabularx}{\linewidth}{c|ccc}
\toprule
Pair matching & ~~copy~~ & ~~DAFNet 0\%~~ & ~~DAFNet 100\%~~ \\
\midrule
expert    & $67_{06}$ & $74_{04}$ & $82_{03}$ \\
automated & n/a        & $71_{06}$ & $80_{03}$ \\
random    & $44_{16}$ & $65_{08}$    & $77_{06}$ \\
\bottomrule
\end{tabularx}
\end{center}
\end{table}

\subsection{Effect of STN} \label{sec:exp_registration}

We assess the need for a registration module with an ablated model. We compare the accuracy of a fused segmentation that is obtained with and without the STN module. Two DAFNet models are compared, trained on ERI data with 100\% and 0\% LGE annotations. The mean Dice without the STN is measured to be $75\pm6\%$ and $71\pm6\%$ respectively. This is lower than the Dice of DAFNet with STN that is $82\pm3\%$ and $74\pm4\%$. Furthermore, in the 100\% case the difference is statistically significant at the 1\% level. Thus, clearly registration helps. 

\begin{figure}[t!]
\centering
\includegraphics[width=\linewidth]{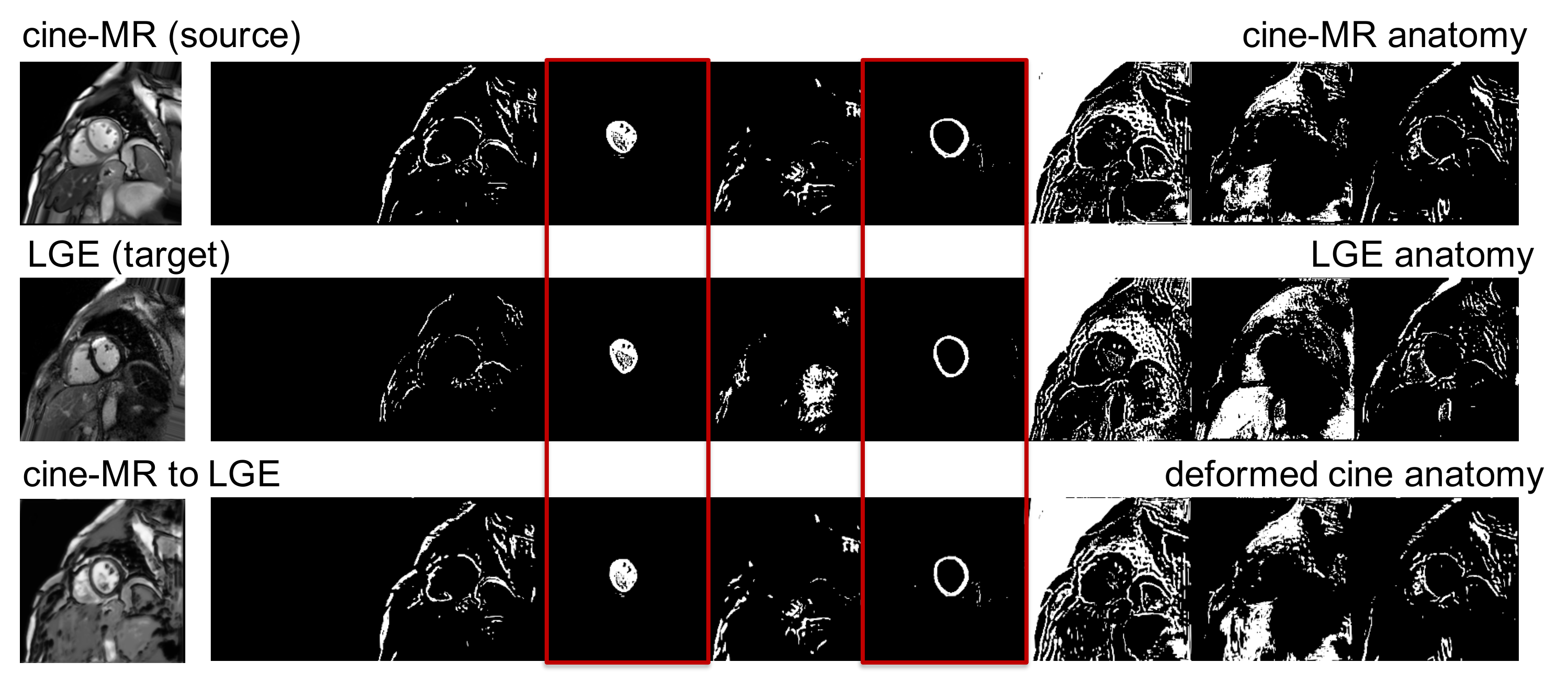}
\caption{Example anatomy alignment. The cine-MR anatomy (row 1) is deformed by the STN to match the LGE one (row 2), resulting in the anatomy of the last row. Red boxes mark channels of areas of interest (left ventricle and myocardium).}
\label{fig:stn_example}
\end{figure}

An example anatomy alignment is shown in Figure~\ref{fig:stn_example}. Although not a perfect alignment of the images is required, the left ventricle and myocardium of the cine-MR have deformed to match the corresponding LGE (marked in red boxes).

\subsection{Ablation study on cost components} \label{sec:exp_costs}

We assess the contribution of cost components in the fused segmentation on ERI with 100\% and 0\% LGE annotations. We evaluate the effect of losses involved in reconstruction, the adversarial training losses on masks ($L_{adv}^{M}$) and images ($L_{adv}^I$), and the modality factor reconstruction loss ($L^z_{rec}$).\footnote{We do not include ablations of the supervised segmentation ($L_{sup}$) because DAFNet is not fully unsupervised.}
Table~\ref{table:ablation} shows that best results are achieved with all cost components. Removing the disentanglement and reconstruction losses significantly reduces performance at 0\% supervision, which conclusively shows their role in learning without annotations.\footnote{We repeat this experiment using a variant without a segmentor, where the anatomy encoder directly predicts segmentations that are aligned by the STN. This performs significantly worse than results of Table~\ref{table:ablation}, showing the importance of an intermediate representation in registration.}
In addition, the data-driven reconstruction cost (via the image discriminator) contributes by encouraging more 
accurate synthesis, which helps learn better anatomical representations.

\begin{table}[t]
\begin{center}
\caption{Ablation study on the effect of individual cost components on LGE segmentation accuracy.}
\label{table:ablation}
\setlength\tabcolsep{5.8pt} % default value: 6pt
\begin{tabularx}{\linewidth}{ccccc|cc}
\toprule
$L_{KL}$ & $L_{rec}$ & $L_{adv}^M$ & $L_{adv}^X$ & $L^z_{rec}$ & DAFNet~0\% & DAFNet~100\% \\
\midrule
--- & --- & \checkmark & --- & --- & $66_{07}$ & $81_{03}$ \\
\checkmark & \checkmark & --- & \checkmark & \checkmark & $72_{02}$ & $80_{03}$ \\
\checkmark & \checkmark & \checkmark & --- & \checkmark & $65_{05}$ & $71_{17}$ \\
\checkmark & \checkmark & \checkmark & \checkmark & --- & $71_{04}$ & $81_{03}$ \\
\checkmark & \checkmark & \checkmark & \checkmark & \checkmark & $74_{04}$ & $82_{03}$ \\
\bottomrule
\end{tabularx}
\end{center}
\end{table}

\subsection{Ablation on factor sizes $C$ and $n_z$}

In all experiments, factor sizes are set to $C$=8, and $n_z$=8.  $C$ is determined experimentally, so that there is enough capacity for all segmentation classes and background anatomy. 
A large $C$ does not affect segmentation, and the redundant capacity is ignored, see ``empty'' channels of Figure~\ref{fig:anatomical_factors}. This is confirmed with ablated models with $C$=4 or $C$=16 trained with 100\% annotations on ERI and CHAOS. The ERI model achieves $82\pm 2\%$ for both setups, the same as when $C$=8. The CHAOS model achieves $74\pm 12\%$ and $85\pm 5\%$ for $C$=4 and $C$=16, respectively. The performance significantly drops when $C$=4, since there is not enough capacity.

Size $n_z$ is determined according to our previous~\cite{chartsias2019disentangled}, and related work~\cite{huang2018multimodal}. We experimented with $n_z$=4 and $n_z$=16 and 100\% annotations on ERI and CHAOS. We find no effect on segmentation accuracy. However, $n_z$ affects the information capacity, approximated by the average variance~\cite{burgess2018understanding}, of each $z$-dimension, where smaller variance implies higher informativeness. With $n_z=16$, the lowest variance is 0.63, the first 8 dimensions have an average of 0.86 and the remaining 8 an average of 0.95. For $n_z$=8, the variance ranges between 0.47 and 0.80, and for $n_z$=4, between 0.43 and 0.60. Admittedly, lower $n_z$ results in higher information content in each dimension, thus large $n_z$ seems redundant in this setup.

\subsection{Effect of decoder design on segmentation accuracy} \label{sec:exp_decoder}

The modular design of DAFNet permits incorporation of components with different designs. We evaluate segmentation accuracy achieved by two decoder architectures: FiLM and SPADE. Specifically, we train a SPADE-based DAFNet on ERI and CHAOS and compare with the FiLM-based DAFNet 
for 100\% and 0\% annotations.

With 100\% annotations, the SPADE-based DAFNet achieves $82\pm3$\% and $85\pm5\%$ on ERI and CHAOS respectively, identical to the Dice achieved by  FiLM. With 0\% annotations, the SPADE-based DAFNet achieves $73\pm4\%$ and $75\pm7\%$, whereas FiLM-based results are $74\pm4\%$ and $74\pm3\%$ respectively on ERI and CHAOS.

We can conclude that the regularising effect of the reconstruction process on extracting segmentations, is similar with both decoder variants. However, different decoder designs influence the way the anatomy and modality factors interact to produce a synthetic image. We explore this next.

\begin{figure}[t]
\centering
\includegraphics[width=\linewidth]{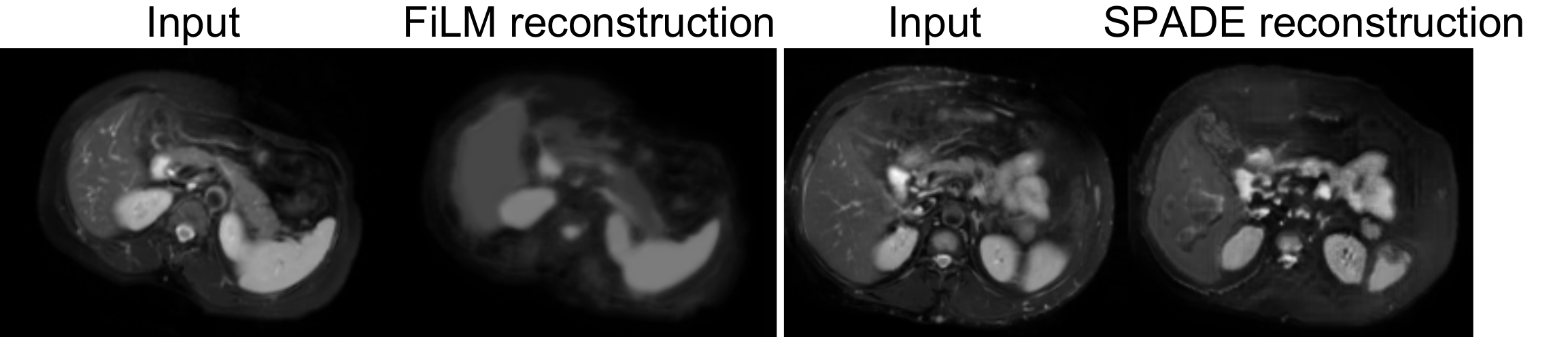}
\caption{Reconstructions with two decoders. The FiLM synthetic image is more flat and lacks texture in contrast to the SPADE synthetic image. Images taken from CHAOS dataset.}
\label{fig:film_vs_spade}
\end{figure}

\subsection{Evaluating disentanglement} \label{sec:exp_disentanglement}
Even though FiLM and SPADE decoders do not result in evident differences in segmentation accuracy, they produce synthetic images of different quality (Figure~\ref{fig:film_vs_spade}). Since the anatomy factors contain flat regions, FiLM-based conditioning with scalar parameters tends to produce images with less texture details than SPADE-based conditioning. 

Here, we assess the information retained in the modality factors and characterise the achieved disentanglement.
This is a challenging problem not addressed in existing literature: all assume vector latent variables (e.g. BetaVAE score~\cite{kingma2013auto}). In DAFNet, and typically in content/style disentanglement, the factors of variation are not of the same dimensionality, with the anatomy being spatial.
For the experiments below, we use models trained on CHAOS with 100\% T2 annotations to assess (dis)entanglement using classification tests, factor arithmetics, and a proposed metric of independence of random variables.

\begin{figure}[t!]
\centering
\includegraphics[width=\linewidth]{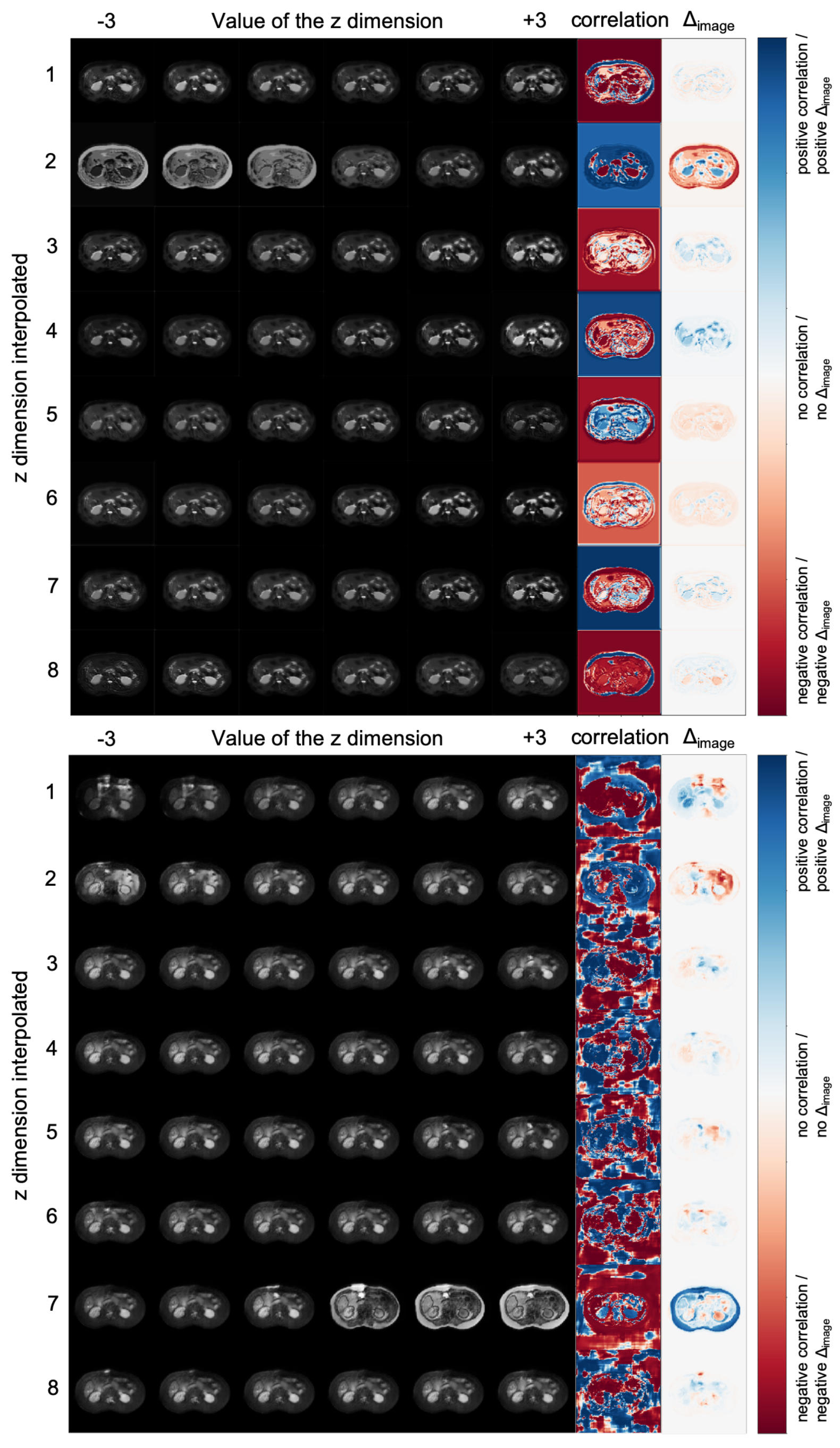}
\caption{(a) FiLM and (b) SPADE based reconstructions. Images per row correspond to interpolating a single $z$ dimension. Last two columns (correlation, and difference image $\Delta_{image}$), indicate regions mostly affected by each $z$ dimension.}
\label{fig:film_panel}
\end{figure}

\subsubsection{Modality classification} \label{sec:disentanglement_modality_classifier}
On the premise that the common modality encoder correctly extracts modality features, a classifier should detect the modality type, given just the $z$-factor. We assess this hypothesis, by training a logistic regression classifier to predict whether different $z$-factors correspond to T1 or T2 images. The classifier's accuracy is 99\% and 97\% for FiLM and SPADE, respectively, on a test set of three subjects. 

We further evaluate whether specific dimensions in $z$ capture the modality type by repeating the experiment for each dimension. In the FiLM model, the 2nd dimension achieves 100\% accuracy, whereas
the rest vary between 54\% and 64\%. Similarly, in the SPADE model the 7th dimension achieves 97\% accuracy vs. 42\% and 63\% of the others.

\subsubsection{Modality factor arithmetics} \label{sec:disentanglement_modalityfactor}
We qualitatively examine the information retained in each dimension of vector $z$ with latent space arithmetics. The likelihood of the modality factor approximates a Gaussian prior, and therefore interpolating in the range $[-3, 3]$ covers the probability space. Figure~\ref{fig:film_panel} shows synthetic images arranged in a grid; images of each row are produced by interpolating the values of a single dimension of $z$, with the remaining ones fixed. The final two columns highlight affected regions by calculating the per-pixel Pearson correlation, as well as the difference, $\Delta_{image}$, between the synthetic images at extreme $z$ values $-3$ and $3$, respectively. 

Both decoders have one $z$-dimension with a global effect ($2$ and $7$ respectively) controlling the ``modality'' type. This finding is inline with the classification results above. Furthermore, some dimensions of the FiLM decoder, e.g. the 8th, appear focused on specific regions, whereas the dimensions of the SPADE decoder produce more diffused correlation images. The latter likely relates to SPADE generating texture better.

\subsubsection{Disentanglement metric} \label{sec:disentanglement_reconstruction}

We propose distance correlation~\cite{szekely2007measuring} as a metric of factor independence, which is invariant to the input variable dimensionality, and can also detect linear and non-linear associations. While it has been used before for reducing data leakage~\cite{vepakomma2019reducing}, we use distance correlation for measuring (dis)entanglement. 
Distance correlation is equal to
\[
dCor(s,z)=\frac{dCov(s,z)}{\sqrt{dVar(s)dVar(z)}},
\]
where $dCov(s,z)$ is the distance covariance of $s$ and $z$, and $dVar(.)$ is the distance variance respectively.
Given $n$ random samples $s_k$ and $z_k$ with $k \in [1, n]$, the distance covariance is the product of two distance matrices (one for each variable) averaged by $n^2$, where each distance matrix $d(.)$ is double centred by subtracting the mean row, the mean column and the overall mean from each element: $dCov^2(s,z)=\frac{1}{n^2}\sum_{i=1}^n \sum_{j=1}^n d(s_i,s_j)d(z_i,z_j)$. The distance variance is then $dVar^2(s)=dCov^2(s,s)$, and $dVar^2(z)=dCov^2(z,z)$.

The distance correlation between $s$ and $z$ values from a FiLM-based model is $dCor=0.55$, whereas the equivalent for a SPADE-based model is $dCor=0.78$. This suggests that factors obtained by the FiLM decoder are more independent and therefore the FiLM-based model is more disentangled. 
Although distance correlation cannot explicitly evaluate the type of information in each variable, this result can be explained intuitively by the decoder design. The SPADE decoder allows more flexibility to the $z$ factors as evidenced in the synthetic images, which contain more texture, but also in the diffused correlation images of Figure~\ref{fig:film_panel}b, implying a higher anatomical correlation (and higher entanglement) between factors.

\section{Conclusion} \label{sec:conclusion}

We have presented a method for multimodal learning and specifically multimodal segmentation, that is robust to the requirement for registered and paired input images. This has been made possible by disentangling images into semantic anatomy factors that are consistently represented across modalities and modality factors that model the intensity variability of the multimodal inputs into a smooth latent space.

We have proposed DAFNet, which, to the best of our knowledge, is the first work that enables multimodal segmentation by aligning disentangled anatomical representations and can be trained with zero annotations for one of the modalities.
We presented the benefit of multimodal (over single-modal) learning in cardiac and abdominal segmentation, where we achieve high accuracy and low variance by fusing anatomical information of different modalities. 
We further demonstrated robustness to misalignments between multimodal images (achieved by a spatial transformer network) and robustness to the quality of the multimodal pair matching (with an optional pair weighting), both made possible by comparing the anatomy factors.
Finally, we made a first step in evaluating the quality of the content/style disentanglement using the distance correlation.

The significance of our work lies in the potential for the use of disentangled representations in other challenging problems of medical research. Future directions include the learning of further factorisations suitable for medical data, capturing pathological information and specific artefacts for instance, as well as a theoretical characterisation of the disentangling process and precise quantification of the type of information that is captured by each factor. Admittedly this is more complex in content/style disentanglement than in  vectorised latent spaces for which metrics have been recently suggested~\cite{do2019theory}.

\bibliographystyle{plain}
\bibliography{references}

\end{document}